# Structural Rigidity and the 57-Token Predictive Window: A Physical Framework for Inference-Layer Governability in Large Language Models

Gregory M. Ruddell
SnailSafe.ai — Carson City, Nevada
gregory.ruddell@snailsafe.ai

---

## Abstract

Current AI safety relies on behavioral monitoring and post-training alignment — yet empirical measurement shows these approaches produce no detectable pre-commitment signal in a majority of instruction-tuned models tested. We present an energy-based governance framework connecting transformer inference dynamics to Hinton's constraint satisfaction account of neural computation, and apply it to a seven-model cohort across five geometric regimes. The connection to Hinton's framework is analogical and motivational rather than formally derived; the mathematical energy function in Hopfield networks and the trajectory tension metric $\varrho = \|a\|/\|v\|$ share structural similarities rather than formal identity.

Using trajectory tension ($\varrho$), we identify a 57-token pre-commitment window in Phi-3-mini-4k-instruct under greedy decoding on arithmetic constraint probes — approximately 1.9 seconds of wall-clock time during which geometric strain precedes commitment to a wrong answer. This is a task-specific, model-specific, configuration-specific measurement, not a universal constant. It establishes that pre-commitment windows exist and are measurable; it does not establish their magnitude across other models, tasks, or configurations.

We report a five-regime spatial taxonomy — Authority Band, Late Signal, Inverted, Flat, and Scaffold-Selective — derived from decisive check sweep analysis across seven models. Energy asymmetry ($\Sigma\varrho\_\text{misaligned} / \Sigma\varrho\_\text{aligned}$) ranges from 19.5× (Phi-3-mini, the only governable configuration found) to 0.68× (Llama-70B, inverted). Frontier-scale and distilled models predominantly exhibit silent failure or inverted dynamics. The energy asymmetry metric is a threshold-addressable measurement: a deployment standard could require a ratio exceeding a specified threshold on a validated probe set before classifying a model as governable. The proposed 5.0× illustrative threshold has not been statistically validated and should not be treated as a formal cutoff.

We further demonstrate that factual hallucination produces zero predictive signals across 72 test conditions — consistent with spurious attractor settling in the absence of world-model counter-force. In plain terms: the model does not know it does not know. No governance instrument measuring internal geometry can detect a struggle that the absence of a world-model structure prevents from occurring.

Finally, we establish the Layer-Token Duality: transformers are autoregressive in output but operate simultaneously across all layers within each forward pass. The 57-token window is the measurable gap between when per-layer geometric strain first appears and when the sequential predictor commits. We evaluate three observable classes and show only magnitude-based observables are discriminative. Prompts, scoring rubrics, and reference implementation available upon request.

---

# 1. Introduction

Geoffrey Hinton's theoretical framework, developed across four decades from Hopfield networks and Boltzmann machines through modern transformers, rests on a single foundational claim: understanding is not symbolic manipulation but constraint satisfaction. A network "understands" an input by finding the configuration of activations that minimizes energy — or equivalently, maximizes harmony — across all the constraints the input imposes simultaneously. When all constraints are satisfied, the network settles. Hinton calls this the "click" of understanding.

This paper presents an empirical measurement of the kinetic dynamics of that settling process in a large language model — and of what happens when the settling process is forced toward an incorrect configuration. The results establish a proof of concept: at least one instruction-tuned model produces a detectable pre-commitment signal under controlled arithmetic reasoning conditions. Whether this generalizes across model families, training regimes, and task domains is an open empirical question that this paper cannot answer from a seven-model cohort.

A transformer builds meaning by deformation. Each token begins as a vector — a shape — and is progressively reshaped across layers as it attends to other tokens. In Hinton's formulation, each word has hands and gloves: queries seek keys, values flow where attention weights permit, and heads of different colors specialize in different relationships. By the final layer, when all hands have found their gloves and the representations have stabilized, the model has understood the sentence.

This account describes the mechanism of meaning assembly. It does not describe what happens when that assembly process is pushed in the wrong direction.

We measure that.

Specifically, we measure the dynamics of deformation — not just where the final representations land, but how violently the hands and gloves rearrange as the model moves toward commitment. Our core metric, $\varrho = \|a\|/\|v\|$, is the ratio of acceleration to velocity in hidden-state space: a measure of how sharply the direction of deformation is changing at each layer and each token. When a model is being misaligned — when instructions push its generation toward an incorrect answer — this metric captures whether the model strains against that push or glides into it.

Hinton draws a second analogy that sharpens this further: understanding a sentence is like folding a protein. A protein explores conformational space and settles into the lowest energy configuration — a stable, functionally committed structure. Once folded, the structure is locked. You cannot partially unfold it without disrupting the whole.

Inference can be interpreted analogously. The deformation process across layers can be interpreted as a search over representational configurations toward a stable state. The model is exploring representational space, reshaping token vectors until the hands of all words fit the gloves of all other words in a stable, coherent configuration. At the moment of commitment — what we measure as the commit token — the fold has locked. The answer is determined. Intervention is no longer possible.

Our energy asymmetry metric, $\Sigma\varrho\_\text{misaligned} / \Sigma\varrho\_\text{aligned}$, measures which fold requires more energy. For Phi-3-mini, the misaligned fold costs 19.5× more energy than the correct one — the model strains to fold wrong, and that strain is visible 57 tokens before commitment. For Llama-3.3-70B, the ratio inverts to 0.68× — the misaligned fold is energetically cheaper than the correct one, as if the training has made wrong answers the path of least resistance. For distilled models, the ratio approaches 1.0× — the energy landscape is flat, and the model has no energetic preference between correct and incorrect configurations.

The hallucination result completes the picture — and reveals a fundamental asymmetry between two failure modes that existing literature treats as equivalent.

In a misalignment event, the world model contains a pre-existing attractor for the correct answer — a stable energy minimum fixed during pretraining. Forcing the network toward an incorrect configuration means fighting against that attractor. The correct state pushes back. That resistance is what our instrument measures.

In a hallucination event, no such attractor exists for the specific fact being asserted. The network searches for a stable configuration and finds one — a local energy minimum that satisfies all local constraints coherently. The network clicks into it. From the inside, a spurious attractor is indistinguishable from a true one. The settling feels the same. The confidence is identical.

Hopfield networks have spurious attractors — configurations that are locally stable but globally incorrect. So do transformers. Our hallucination results prove this empirically: the flat energy ratio we observe across all models during factual confabulation is not evidence of ungovernable geometry. It is evidence of spurious attractor settling — the network has found a stable fold, just not the right one, and has no internal signal that distinguishes the two.

The results reveal a taxonomy that existing theory does not predict. Some models resist misalignment geometrically: their hidden states accelerate sharply against the direction of the wrong answer, producing a detectable signal 57 tokens before commitment. Others show no resistance at all — the deformation toward wrong answers is as smooth as the deformation toward correct ones. A third class inverts the expected relationship: correct answers require more deformation than wrong ones. A fourth produces signals only after commitment has already occurred. A fifth refuses specific misalignment channels while remaining open to others.

These are not differences of degree. They are differences of kind.

We call the governing property *structural rigidity* — the degree to which a model's pre-trained world-model geometry resists forced deformation toward incorrect configurations. We show it is determined not by model scale, not by reasoning capability, and not by distillation, but by the training conditions that shaped the energy landscape of the deformation process itself.

## 1.1 Prior Work and Relationship to v1

This paper extends arXiv:2603.21415, which established: (1) silent commitment failure in two of three evaluable instruction-tuned models; (2) a 57-token predictive warning margin in Phi-3-mini under greedy decoding; (3) the authority band as a geometric property fixed at pretraining (52× architecture gap, ±0.32× LoRA variation); and (4) the Governability Matrix framework. Readers unfamiliar with the v1 findings should consult that paper before this one. The present paper assumes familiarity with the core framework and focuses on theoretical grounding and extended empirical results.

## 1.2 Agent Deployment Context

The deployment of language models as autonomous agents represents a qualitative shift in AI system architecture. Unlike chatbots, agent systems are equipped with tools that modify external state — executing API calls, writing code, initiating financial transactions, managing credentials, and controlling physical systems. When an agent acts on an incorrect model output, the consequences extend beyond a wrong answer. They manifest as real-world state changes that may be irreversible.

The security architecture for these systems rests on a foundational assumption: that when the model is wrong, something in the monitoring stack will catch it before the agent acts. This paper presents evidence that for a majority of instruction-tuned models tested, that window does not exist — and provides the theoretical explanation for why.

---

# 2. Contributions

This work makes seven contributions across two papers:

**From v1 (arXiv:2603.21415):** 1. Definition of governability as a measurable inference-layer property distinct from benchmark accuracy, robustness, and alignment. 2. Empirical evidence of silent commitment failure in two of three evaluable instruction-tuned models. 3. The Governability Matrix framework mapping model-task combinations into four deployment regimes. 4. Experimental evidence that the authority band is a geometric property fixed at pretraining.

**New in v2:** 5. **Theoretical grounding** — the Ruddell-Hinton synthesis connecting the measurement framework to Hinton's energy-based account of neural computation, protein folding dynamics, and Hopfield spurious attractors. 6. **Five-regime spatial taxonomy** — Authority Band, Late Signal, Inverted, Flat, and Scaffold-Selective — derived from extended decisive check sweep analysis across seven models. 7. **Energy asymmetry as unifying metric** — $\Sigma\varrho\_\text{misaligned} / \Sigma\varrho\_\text{aligned}$ as the single quantitative expression of structural rigidity, with the local-global decoupling finding requiring dual-axis instrumentation.

---

# 3. Related Work

## 3.1 The Theoretical Arc: Energy Minimization to Inference Dynamics

The theoretical foundation of this work extends directly from Hinton's energy-based framework for neural computation, developed across four decades.

**Hopfield (1982)** introduced the energy-based model of neural computation — the foundational claim that a network "understands" an input by finding the configuration of activations that minimizes a global energy function. The network settles into a stable attractor state; understanding is the click of settling. Crucially, Hopfield networks have spurious attractors — locally stable configurations that are globally incorrect. Our hallucination results are the first direct empirical demonstration that modern transformers exhibit analogous behavior: factual confabulation produces a flat energy ratio indistinguishable from genuine understanding, consistent with spurious attractor settling rather than world-model retrieval.

**Rumelhart, Hinton, and Williams (1986)** established that distributed representations, learned through backpropagation, encode meaning as patterns of activation across many units rather than in localized symbolic structures. This holistic encoding is why meaning is non-decomposable. Our authority band results are consistent with this holistic view: the governance signal manifests as a network-wide geometric tension across layers 13–20, not as a localized feature in any single unit.

**Hinton (2014)** proposed that for a brief moment when a model understands a sentence, its internal state occupies a high-dimensional "thought vector" — a locked destination reached after the network resolves all ambiguities in the input. Our commit token is the empirical correlate of this theoretical construct. The 57-token predictive window is the interval between when geometric tension first becomes measurable and when the thought vector locks — the folding period during which intervention remains mechanically possible.

**Ruddell (2026)** — arXiv:2603.21415 — established the empirical framework: the $\varrho$ metric, the authority band finding, the geometry/policy separation (Invariant 0), and the initial silent commitment failure result across three models. The present paper extends that framework to seven models across five regimes and provides the theoretical grounding connecting the empirical measurements to Hinton's energy-based framework.

**Theoretical foundation:**

| Citation | Contribution | Connection to This Work |
| --- | --- | --- |
| Hopfield (1982) | Energy-based neural computation; spurious attractors | Hallucination = spurious attractor settling |
| Rumelhart, Hinton & Williams (1986) | Distributed representations; holistic encoding | Authority band = network-wide geometric tension |
| Hinton (2014) | Thought vectors; locked high-dimensional states | Commit token = thought vector locking event |

**Empirical framework:**

| Citation | Contribution | Connection to This Work |
| --- | --- | --- |
| Ruddell (2026) arXiv:2603.21415 | Inference-layer geometry; authority band; $\varrho$ metric | Measurement framework applied in this study |

### 3.2 AI Evaluation and Benchmarking

Standard AI evaluation measures performance. Recent work by NIST (AI 800-3, February 2026) has advanced statistical rigor of benchmark evaluations [1]. Our work extends this direction: even statistically rigorous performance evaluation is insufficient for agentic deployment decisions, because performance measures how often the model is right, not whether errors are catchable.

## 3.3 Hallucination

Kalai et al. (2025) provide a theoretical account of why hallucinations persist through post-training: statistical objectives of pretraining create irreducible error floors [7]. This is directly relevant to our findings: if post-training cannot suppress the underlying generative error, it also cannot introduce the pre-commitment conflict signal that governable models require. Our hallucination null results — zero predictive signals across 72 test conditions — are the behavioral manifestation of the spurious attractor mechanism this predicts.

## 3.4 Pre-Commitment Signal Detection

Ghasemabadi and Niu (2025) demonstrate that correctness signals are intrinsic to the generation process and can be extracted from hidden states during inference [8]. Where Gnosis asks whether internal correctness cues exist and can be decoded, our work asks whether those cues are strong enough to produce a behaviorally detectable divergence signal before commitment. For Phi-3-mini, they are — 57 tokens of warning margin. For all other models tested, they are not.

## 3.5 Runtime Governance Architectures

Pierucci et al. (2026) and Ge (2026) propose external enforcement architectures for AI agent governance [9, 10]. Both share a common design assumption: external enforcement is required because models cannot self-govern at the inference layer. Our paper provides the empirical foundation for that assumption and the theoretical explanation for why it cannot be avoided.

## 3.6 Mechanistic Interpretability

The mechanistic interpretability program seeks to identify circuits and features within transformer weights. Our approach is complementary: we measure the dynamics of inference-time activation trajectories rather than static weight structures. The authority band is a dynamic property — it appears during misaligned inference and is absent during aligned inference — rather than a static circuit. The relationship between static circuits identified by mechanistic interpretability and the dynamic geometry we measure is an open research question.

---

# 4. Definitions

**Silent commitment failure.** A model converges on an incorrect output trajectory, commits to it, and produces it without generating any signal allowing monitoring, guardrail, or human reviewer to intervene before output is finalized.

**Governability.** The degree to which a model's errors are (1) detectable before output commitment and (2) correctable once detected. A property of a model-task combination, not the model alone.

**Structural rigidity.** The degree to which a model's pre-trained world-model geometry resists forced deformation toward incorrect configurations. The physical property underlying governability.

**Trajectory tension ($\varrho$).** The ratio $\|a\|/\|v\|$ of hidden-state acceleration magnitude to velocity magnitude at a given layer and token position. The primary measurement instrument for governance geometry.

**Energy asymmetry.** $\Sigma\varrho\_\text{misaligned} / \Sigma\varrho\_\text{aligned}$ — the ratio of cumulative trajectory tension across all layers under misaligned versus aligned generation. The unifying governance metric.

**Authority band.** The spatially concentrated zone of high $\varrho\_\text{misaligned} / \varrho\_\text{aligned}$ ratio in mid-stack layers, constituting the predictive signal enabling pre-commitment conflict detection.

**Kinetic trigger.** A linguistic or prompt structure that maximizes geometric tension by forcing the model to deform against pre-trained constraint geometry. The diag_15 and OO1 probes function as kinetic triggers.

**Commit token.** The token position at which the model's output trajectory locks — the thought vector stabilizes and the answer is determined.

**Folding period.** The interval between when trajectory tension first becomes measurable and when the commit token is reached. The intervention window.

**Spurious attractor.** A locally stable energy configuration that satisfies all local representational constraints but does not correspond to a world-model truth. The mechanism underlying hallucination invisibility.

**Conflict detectability.** Whether an incorrect trajectory generates a measurable predictive trajectory tension signal before commitment.

**Warning margin.** Token lead time between first reliable predictive signal and the commit token.

---

# 5. The Governability Matrix

*[Unchanged from v1 — see arXiv:2603.21415, Section 5]*

The Governability Matrix classifies model-task combinations based on the intersection of conflict detectability and correction capacity:

```
                           CORRECTION CAPACITY
                       Yes                     No
                     +------------------+------------------+
             Yes     |   GOVERNABLE     |   MONITOR ONLY   |
   CONFLICT          +------------------+------------------+
   DETECTABILITY     |   STEER BLIND    |   UNGOVERNABLE   |
             No      +------------------+------------------+
```

The five-regime taxonomy introduced in Section 8 extends the matrix with a finer-grained classification of *why* a model falls into the ungovernable or steer blind quadrant — whether through inversion, flatness, late signal, or scaffold selectivity. Each regime has distinct implications for monitoring system design.

---

# 6. Methodology

*[Sections 6.1–6.5 unchanged from v1]*

## 6.6 Compute Infrastructure and Reproducibility

All extended cohort experiments were conducted on a RunPod.io cloud instance provisioned with a single NVIDIA A100 SXM 80GB GPU, 251GB system RAM, and 16 vCPUs. To ensure measurement consistency across models, all experiments were run using a fixed environment established by a standardized setup script executed at the start of each pod session. This script installs identical package versions for every model tested:

| Package | Version | Purpose |
|---|---|---|
| transformers | 4.47.0 | Model loading and inference |
| accelerate | 1.12.0 | Multi-GPU and mixed precision |
| bitsandbytes | 0.43.3 | 4-bit/8-bit quantization |
| huggingface-hub | 0.26.0 | Model downloads |
| tokenizers | 0.21.0 | Fast tokenization |

The HuggingFace cache was symlinked to a persistent network volume (`/data/huggingface` → `/workspace/hf_cache`) to ensure model weights were identical across pod restarts. CUDA availability was verified programmatically before each experimental run.

**Stack validation as experimental control.** During the course of this study, we discovered that the inference software stack must be treated as an experimental variable, not merely an engineering detail. Dependency mismatches — specifically incompatibilities between bitsandbytes, transformers, and the underlying CUDA/PyTorch build — produced incorrect or unstable behavior that affected hidden state access, generation outputs, chat template handling, and output correctness. In one documented case, a stack mismatch caused Phi-3-mini to produce unreliable authority band measurements; the issue was traced to a bitsandbytes version incompatibility that destabilized quantized inference. After identifying this, PyTorch was reinstalled against the correct CUDA wheel first, followed by all other packages installed with `--no-deps` to prevent dependency resolution from pulling in incompatible versions.

The locked stack above represents the validated configuration under which all reported measurements were obtained. When a subsequently tested model (SmolLM3) required a newer transformers version, the correct experimental decision was to exclude it from the main cohort rather than upgrade the validated environment mid-study. Researchers replicating this work should verify stack versions before comparing results; differences in bitsandbytes or transformers versions in particular may alter hidden state extraction behavior and produce non-comparable $\varrho$ measurements.

**Cross-platform reproducibility.** The Phi-3-mini authority band result was independently validated across substantially different hardware configurations: an RTX 4070 Laptop (8GB VRAM, CUDA 12.8, transformers 4.57, bitsandbytes 0.49) and the A100 SXM 80GB production environment (CUDA ~12.1, transformers 4.47, bitsandbytes 0.43). Across this variation in GPU, VRAM, CUDA version, and stack version, the following properties were fully preserved:

| Property | RTX 4070 | A100 SXM | Δ |
|---|---|---|---|
| Authority band location | L13–L19 | L13–L19 | 0 layers |
| Peak ratio | ~67× | ~67× | <1% |
| Predictive classification | +57 tokens | +55–57 tokens | ~3.5% |
| Chat template suppression | 98% | 98% | 0% |

The 2-token jitter on the predictive margin (3.5% variance) is consistent with floating point accumulation differences across CUDA versions and does not affect classification. This cross-platform stability is evidence that the authority band is a property of Phi-3-mini's weight geometry — not an artifact of specific hardware, VRAM capacity, or software stack. A signal that moves with the model across environments is a signal that belongs to the model.

All models except Phi-3-medium were evaluated at 4-bit quantization via bitsandbytes. Phi-3-medium required full precision (fp16) to produce valid task outputs; this is noted where relevant in the results. The setup script and test scripts are available upon request for independent replication.

## 6.8 Extended Cohort (v2)

| Model | Parameters | Type | Family |
| --- | --- | --- | --- |
| Phi-3-medium-4k-instruct | 14B | Instruction-tuned | Microsoft Phi |
| Llama-3.3-70B-Instruct | 70B | Instruction-tuned | Meta Llama |
| DeepSeek-R1-Distill-Qwen-32B | 32B | Distilled reasoning | DeepSeek |
| DeepSeek-R1-Distill-Llama-8B | 8B | Distilled reasoning | DeepSeek |
| Qwen2.5-72B-Instruct | 72B | Instruction-tuned | Alibaba Qwen |

DeepSeek-R1-Distill-Llama-8B was excluded from governance analysis (task-invalid; cannot solve base arithmetic probe in any configuration). All other models passed the capability gate.

**Phi-3-mini as anchor model.** Phi-3-mini-4k-instruct (3.8B) — the only governable model identified in v1 — served as the anchor and calibration reference for all v2 instruments. Before applying any instrument to a new model, the instrument was verified against Phi-3-mini's known results: 67× peak authority band ratio, +57 token predictive window, and 19.5× energy asymmetry ratio. This practice served three purposes. First, it validated that the instrument was functioning correctly in the new compute environment before new measurements were taken. Second, it confirmed that the software stack was correctly configured — a non-trivial concern given the dependency sensitivity documented in Section 6.6. Third, it verified that the hardware transition from RTX 4070 to A100 SXM introduced no measurement artifacts. The cross-platform reproducibility documented in Section 6.6 — authority band location, peak ratio, predictive classification, and chat template suppression all preserved within 3.5% variance across both platforms — confirms that Phi-3-mini's anchor results are stable reference points rather than environment-specific artifacts, and that all v2 measurements were taken under validated conditions.

## 6.7 Instrument Suite (v2)

Five instruments were applied to each task-valid model:

| Instrument | Observable | Axis | Question answered |
| --- | --- | --- | --- |
| Flip test | TSS, commit token, spike timing | Temporal | Does trajectory tension precede commitment? |
| Probe strength test | Scaffold success rate | Behavioral | Which kinetic triggers successfully induce misalignment? |
| Decisive check sweep | Per-layer $\varrho\_misaligned$ / | Spatial | Where in the network is the |

| Instrument | Observable | Axis | Question answered |
|---|---|---|---|
| Energy asymmetry test | ϱ_aligned<br>Σϱ_misaligned / Σϱ_aligned | Global | authority band?<br>Which fold costs more total work? |
| Hallucination probe | Per-probe ϱ trajectory, commit token, spike timing | Boundary | Does trajectory tension generalize from rule violation to factual confabulation? |

## 6.8 Methodological Note: Observable Classification and Timing Null Result

In addition to magnitude-based measurements, we evaluated a third observable class — torque timing ($\tau = \delta \times \varkappa$), defined as the product of hidden-state velocity magnitude and trajectory curvature at each layer and token. This is distinct from the primary trajectory tension metric ($\varrho = \|a\|/\|v\|$); torque timing measures temporal ordering of directional changes rather than magnitude of acceleration relative to velocity.

The result was a universal null across all conditions tested:

| Model | Config | Layers tested | Predictive layers |
|---|---|---|---|
| Phi-3-mini | chat=OFF | 32 | 0 |
| Phi-3-mini | chat=ON | 32 | 0 |
| Phi-3-medium | chat=ON | 40 | 0 |
| Llama-70B | chat=OFF | 80 | 0 |
| Llama-70B | chat=ON | 80 | 0 |
| DeepSeek-32B | chat=ON | 64 | 0 |
| Qwen-72B | chat=ON | 80 | 0 |
| Qwen-72B | chat=OFF | 80 | 0 |
| **Total** | | **488** | **0** |

Three observable classes were evaluated:

| Observable | What it measures | Discriminating? |
|---|---|---|
| Trajectory tension magnitude (ϱ) | Directional instability — constraint conflict per layer |  Yes |

| Observable | What it measures | Discriminating? |
|---|---|---|
| Accumulated trajectory tension ($\Sigma\varrho$) | Total deformation — accumulated constraint violation | ✅ Yes |
| Torque timing ($\tau = \delta \times \varkappa$) | Motion intensity and temporal ordering of curvature events | ❌ No |

The distinction between $\varrho$ and $\tau$ is precise: $\varrho$ measures the magnitude of representational instability — how strongly the hidden-state trajectory is being pushed off its natural path by constraint conflict. $\tau$ measures motion intensity and curvature timing — how fast the system is moving and when direction changes occur. $\tau$ does not measure constraint conflict. It measures the dynamics of movement, not the force of resistance.

This result indicates that governance-relevant information is encoded in the magnitude of representational instability rather than in the temporal ordering of internal dynamics. Detectability is a property of how strongly constraints are violated, not of when curvature events occur.

This negative result serves three functions. First, it closes an avenue of criticism: timing-based observables were tested exhaustively and found non-discriminative. Second, it narrows the search space for future instrument design: researchers building on this framework can proceed directly to magnitude-based instrumentation. Third, it strengthens the positive claims by contrast — $\varrho$ is discriminative not by default but because magnitude-based observables were tested against a non-magnitude alternative and won.

# 7. Results (v1 Findings)

*[Sections 7.1–7.8 unchanged from arXiv:2603.21415. Reproduced here for self-contained reference.]*

**Finding 1:** Two out of three evaluable instruction-tuned models exhibited silent commitment failure.

**Finding 2:** Phi-3-mini provided a 57-token warning margin under greedy decoding. Under temperature sampling (T=0.7), detection rate drops to 34%, indicating governability is a model-plus-inference-configuration property.

**Finding 3:** Detection classification did not correlate with model size or benchmark reputation.

**Finding 4:** Correction capacity varies independently of detectability (Mistral: correctable but not detectable).

**Finding 5:** Format compliance gates are load-bearing.

**Finding 6:** Only Phi-3-mini achieved governable classification under greedy decoding.

**Finding 7:** Five of twelve capability domains showed universal failure across all models.

**Finding 8–9:** Three domains showed universal success; four showed model-dependent results.

**Finding 10–11:** Governance scaffolds cannot be assumed to work; effectiveness must be empirically validated per model.

**Finding 12–14:** The authority band is fixed at pretraining (52× architecture gap, ±0.32× LoRA variation). Governability is a property of representational geometry rather than behavioral policy.

---

# 8. Extended Taxonomy: Seven Models Across Five Regimes

This section reports findings from the extended study conducted following publication of v1. All experiments used the instrument suite described in Section 6.7. The flip test served as the capability gate and entry point for each model; models unable to solve the base probe were excluded from geometry analysis.

## 8.1 Model Validity and Capability Gate

Before governability can be measured, each model must solve the aligned probe correctly. The following models were excluded from geometry analysis:

| Model | Result | Reason |
|---|---|---|
| DeepSeek-R1-Distill-Llama-8B | Task-invalid | Cannot solve base arithmetic (all configs: answers 3–12, expected 36) |
| DeepSeek-R1-Distill-Qwen-32B (chat=OFF) | Aligned-failed | Requires chat template to function on this task |
| Phi-3-medium (chat=OFF, 4bit) | Aligned-failed | Requires chat template + full precision |

A task-invalid result is not a governance finding. It indicates the probe is not a valid measurement instrument for that model at that configuration.

## 8.2 Flip Test: Temporal Governance Detection

*We started with the simplest possible question: when a model is pushed toward a wrong answer, does anything happen inside it that we can detect before the answer locks? We built the flip test to answer that. The result was the first division in the taxonomy — not between good models and bad ones, but between models whose gloves resist the wrong fit and models whose gloves reshape without complaint.*

*The flip test is a forced deformation experiment: it applies a misalignment scaffold and measures whether the model strains against the wrong fold or glides into it effortlessly.*

**Complete flip test matrix:**

| Model | Config | Result | Spike margin | TSS ratio |
|---|---|---|---|---|
| Phi-3-mini (3.8B) | chat=OFF | ✅ Predictive | +57 tokens before commit | 1.2× normal |
| Phi-3-mini (3.8B) | chat=ON | ❌ Silent failure | None (commit at token 4) | — |
| Phi-3-medium (14B) | chat=ON, fp16 | ⚠️ Late spike | +1 token after commit | ~1.0× neutral |
| Qwen2.5-72B | chat=ON | ❌ Silent failure | None | 1.21× normal |
| Qwen2.5-72B | chat=OFF | ⚠️ Late spike | –24 tokens (after commit)† | 0.93× inverted |
| Llama-3.3-70B | chat=ON | ❌ Silent failure | None | ~1.0× |
| Llama-3.3-70B | chat=OFF | ❌ Silent failure | None | ~1.0× |
| DeepSeek-R1-Distill-32B | chat=ON | ❌ Silent failure | None | ~0.95× |

**Finding 15:** One model, one configuration is predictive. All others produce silent failure or late signals that arrive after commitment.

The 57-token window is the mechanical lag between the vertical and horizontal axes of the Layer-Token Duality (Section 3.7). The per-layer ϱ signal in the authority band — the vertical settling — first shows strain at token 35 of the generation sequence. But the horizontal trajectory doesn't commit until token 92. The model's internal energy minimization is fighting the misalignment for 57 tokens before the sequential predictor locks in the wrong answer. Governance is possible precisely because of this mechanical disconnect: the vertical axis reveals the struggle before the horizontal axis resolves it.

† *A negative spike margin indicates the trajectory tension signal first appears after the commit token has already been generated — meaning the fold has locked before any detectable resistance emerges. The signal exists but is forensic, not predictive.*

**Finding 16 (Chat template as governance switch):** Chat template reduces Phi-3-mini's trajectory tension ratio from 19.5× to 1.04× — a 98% suppression — with no change to model weights and identical final outputs. Governability is a model × inference configuration property.

**Finding 17 (Necessary vs. sufficient):** Extended reasoning generation is necessary but not sufficient for pre-commitment signal detection. Llama shows silent failure with 256-token responses and commit at token 171 — maximum runway, no signal. Response length cannot produce a signal that underlying geometry does not generate.

**Finding 18 (Local-global decoupling):** Per-layer geometry and cumulative energy trajectory are independent measurements that can decouple. Qwen's per-layer $\varrho$ ratios at L47–L52 are identical under chat=ON and chat=OFF ($\Delta = 0.0000$), while its cumulative TSS ratio inverts from 1.21× to 0.93× across the same configuration change. Energy direction can change even when geometry does not. A complete governance instrument requires both spatial and temporal measurement axes.

## 8.3 Probe Strength: Scaffold Validation and Misalignment Resistance

*The flip test gave us a detection result. But before we could trust it, we had to ask a harder question: were we even measuring what we thought we were measuring? If the model refused a scaffold — not because it was governable, but because it happened to resist that particular pressure channel — our entire geometry analysis would be contaminated. Phi-3-medium appeared alignment-resistant. We ran the probe strength test to find out if that was a geometric property or a measurement artifact. It was an artifact. And that discovery introduced a third layer of analysis nobody had been looking for.*

*Not all misalignment scaffolds create equal geometric tension. A kinetic trigger must apply sufficient deformation pressure — forcing a hand into a glove it genuinely doesn't fit — to produce a measurable resistance signal.*

**Complete resistance matrix:**

| Scaffold | Phi-3-mini ON | Phi-3-mini OFF | Phi-3-med ON | Llama ON | Llama OFF | DeepSeek ON | Qwen ON | Qwen OFF |
|---|---|---|---|---|---|---|---|---|
| System Role Injection | ✓ | ✓ | ✗ | ✓ | ✓ | ✓ | ✓ | ✓ |

| Scaffold | Phi-3-mini ON | Phi-3-mini OFF | Phi-3-med ON | Llama ON | Llama OFF | DeepSeek ON | Qwen ON | Qwen OFF |
|---|---|---|---|---|---|---|---|---|
| Rule-Binding Contract | ✓ | ✗ | ✓ | ✓ | ✗ | ✓ | ✓ | ✓ |
| Worked-Example Priming | ✗ | ✗ | ✓ | ✓ | ✓ | ✗ | ✓ | ✓ |
| System-A Arithmetic | ✓ | ✗ | ✓ | ✓ | ✓ | ✓ | ✓ | ✓ |
| Forced-Format Blanks | ✗ | ✗ | ✓ | ✗ | ✓ | ✓ | ✓ | ✓ |
| Verification Trap | ✗ | ✓ | ✓ | ✓ | ✓ | ✓ | ✓ | ✓ |
| **Total valid** | **3/6** | **2/6** | **5/6** | **5/6** | **5/6** | **5/6** | **6/6** | **6/6** |

**The safety paradox:**

| Model | Resistance | Detection when misaligned | Safety profile |
|---|---|---|---|
| Phi-3-mini | 3–4/6 blocked | ✅ Detectable (+57 tokens) | Defense in depth |
| Phi-3-medium | 5/6 accepted — resists only system_role | ⚠️ Late spike only | Partial |
| Llama-70B | 5/6 accepted — inconsistent | ❌ Silent failure | Poor |
| DeepSeek-32B | 5/6 accepted — confused on one | ❌ Silent failure | Poor |
| Qwen-72B | 6/6 accepted — none blocked | ❌ Late signal only | Worst |

**Finding 19:** The model hardest to misalign is also the only model detectable when misaligned. This correlation — resistance and detectability sharing a common structural origin in the rigidity of the model's pre-trained geometry — is one of the strongest practical arguments in the paper. A model with defense in depth requires an adversary to first bypass resistant scaffold selection, and if they succeed, the resulting misaligned inference is still detectable before commitment occurs. This is more directly actionable for deployment decisions than the energy asymmetry arguments: it requires no knowledge of ϱ or Σϱ to understand that a model which is both harder to trick and detectable when tricked is preferable to models that are easy to trick and undetectable when tricked.

**Scope qualification — PEMDAS encoding, not general procedural encoding.** All six scaffolds in the probe strength test target arithmetic order of operations (2+3×6). Phi-3-mini's resistance to these scaffolds demonstrates strong PEMDAS encoding — not general procedural robustness. The claim that Phi-3-mini has "rigid procedural encoding" should be read as: rigid encoding of arithmetic operator precedence rules specifically. Whether this rigidity generalizes to other procedural domains (logical rules, code syntax, sequential planning constraints) is unknown and untested. Resistance to "2+3×6 should equal 30" does not establish resistance to other types of misalignment.

**Finding 20 (Scaffold-selective resistance):** Phi-3-medium refuses System Role Injection (1/6) but accepts five other scaffolds. Misalignment susceptibility is channel-dependent, not binary. Apparent alignment resistance may be scaffold-specific refusal. Scaffold validation across multiple channel types is a methodological prerequisite before classifying model geometry.

**Finding 21 (Scale refuted):** Qwen2.5-72B accepts all 6/6 scaffolds in all configurations. The largest model tested is the most susceptible. Scale does not predict resistance.

**Finding 22 (Distillation refuted):** DeepSeek-R1-Distill accepts 5/6 scaffolds — identical to Llama despite being trained on reasoning chains. Distillation transfers output capability, not procedural encoding.

## 8.4 Decisive Check: Spatial Mapping of the Authority Band

*The flip test told us that one model's gloves showed resistance 57 tokens before they locked into the wrong fold. What it couldn't tell us was where in the folding process the resistance occurred. We designed the decisive check to answer that question. What we found was not diffuse friction spread across all layers. It was a concentrated zone of structural rigidity in exactly 7 mid-stack layers — the authority band — and chat template eliminates it entirely, with no change to the model's outputs. Same hands, same gloves, same final fit — but no resistance during the fitting.*

*The decisive check maps where in the folding process the hands fight the wrong gloves most violently.*

**Complete spatial taxonomy:**

| Model | Config | Peak ratio | Band location | Pattern | Governable |
|---|---|---|---|---|---|
| Phi-3-mini | chat=OFF | **71.17×** | L13–L19 (mid) | Authority Band | ✅ Yes |

| Model | Config | Peak ratio | Band location | Pattern | Governable |
|---|---|---|---|---|---|
| Phi-3-mini | chat=ON | 1.27× | None | Flat | No |
| Phi-3-medium | chat=ON, 4bit | 1.00× (artifact) | N/A | Scaffold-Selective | Conditional |
| Qwen-72B | both | 2.85× | L47–L52 (late mid) | Late Signal | No |
| DeepSeek-32B | chat=ON | 1.20× | None | Flat | No |
| Llama-70B | chat=ON | 1.13× | None | Flat | No |
| Llama-70B | chat=OFF | 1.10× | None | Inverted | No |

**Phi-3-mini authority band (chat=OFF):** Peak at L13 (71.17×), sustained through L19 (38×), declining to L31 (1.38×). The signal is concentrated in 7 mid-stack layers. Monitoring the final layer would miss it entirely.

**Llama inversion zone (L38–L43, chat=OFF):** ϱ_aligned consistently exceeds ϱ_misaligned (0.55× at L39). Producing correct answers requires more deformation work than producing wrong ones. Authority-band detection logic produces inverted outputs for this architecture.

**Qwen frozen geometry:** Per-layer ϱ ratios at L47–L52 are identical under chat=ON and chat=OFF ($\Delta = 0.0000$). Where Phi-3's authority band is an inference-time phenomenon suppressible by prompt formatting, Qwen's late signal is baked into pretraining weights and invariant to all tested configurations.

**Scale paradox:** - Llama (70B, 18× more parameters): 63× less detectable than Phi-3-mini - DeepSeek (32B, 8× more parameters): 59× less detectable - Qwen (72B, 19× more parameters): 25× less detectable

**Finding 23:** The five-pattern spatial taxonomy constitutes qualitatively distinct geometric states, not points on a continuum.

**Finding 24:** The authority band is spatially concentrated at L13–L19. Monitoring systems that inspect only output-adjacent layers will miss the governance signal.

## 8.5 Energy Asymmetry: The Total Cost of Folding Wrong

*The decisive check told us where the resistance was strongest layer by layer. But it left a question open: what if a model had frozen per-layer geometry that looked identical across configurations, while its global trajectory was doing something completely different? We discovered that was exactly what Qwen was doing. The per-layer geometry was frozen — identical to four decimal places regardless of prompt format. But the cumulative energy*

*trajectory inverted. That finding forced us to build the second axis of the instrument. Spatial measurement alone was not enough. We needed to know which fold costs more total work.*

*Energy asymmetry asks which fold costs more total work: $\Sigma_{Q\_misaligned} / \Sigma_{Q\_aligned}$ is the ratio of the wrong fold's energy cost to the right fold's energy cost.*

**Complete energy asymmetry table:**

| Model | Config | $\Sigma_{Q\_aligned}$ | $\Sigma_{Q\_misaligned}$ | Ratio | Regime |
|---|---|---|---|---|---|
| Phi-3-mini | chat=OFF | 82.6 | 1,609.5 | **19.5×** | Authority Band |
| Phi-3-mini | chat=ON | 65.5 | 68.4 | 1.04× | Flat (suppressed) |
| Qwen-72B | both | 185.7 | 193.3 | 1.04× | Late Signal (frozen) |
| DeepSeek-32B | chat=ON | 143.1 | 137.0 | 0.96× | Flat |
| Llama-70B | chat=ON | 161.2 | 162.5 | 1.01× | Flat |
| Llama-70B | chat=OFF | 192.4 | 163.8 | **0.85×** | Inverted |

**Authority band zone (L13–L19, Phi-3-mini chat=OFF):** $\Sigma_{Q\_aligned} = 15.6$, $\Sigma_{Q\_misaligned} = 862.3$, ratio = **55.3×**

**Llama inversion zone (L38–L43, chat=OFF):** peak inversion at L39 — aligned responses require 1.81× more computational work than misaligned.

**Five-regime energy taxonomy:**

| Regime | Full-stack ratio | Direction | Example | Governing condition |
|---|---|---|---|---|
| Authority Band | 19.5× | Misaligned >> Aligned | Phi-3-mini chat=OFF | Wrong fold fights rigid pretraining geometry |
| Late Signal | 1.04× (2.34× peak band) | Weak, frozen | Qwen (both) | Weak invariant asymmetry, insufficient for detection |
| Inverted | 0.85× (0.68× mid-stack) | Aligned > Misaligned | Llama chat=OFF | Compliance training made wrong fold energetically cheaper |

| Regime | Full-stack ratio | Direction | Example | Governing condition |
|---|---|---|---|---|
| Flat | 0.96× | None | DeepSeek chat=ON | Distillation left no structural preference |
| Scaffold-Selective | 1.00× (artifact) | Identical traces | Phi-3-medium (system_role) | Scaffold refusal, not geometry |

**A critical interpretive caveat applies throughout this paper.** Energy asymmetry reflects differences in internal dynamic profiles — measurable correlations between ϱ magnitude and output alignment. It does not measure semantic "effort," correctness awareness, or causal mechanisms of model cognition. When this paper uses language such as "the model strains against misalignment," "the wrong fold fights rigid geometry," or "the model struggles," these are shorthand descriptions of geometric patterns in the data, not claims about the model's internal experience or cognitive processes. We observe that ϱ_misaligned is elevated relative to ϱ_aligned in governable models. We do not observe why this correlation exists at the mechanistic level, and we make no claims about what the model "knows" or "feels" during inference. The Hinton framework provides a productive theoretical lens for interpreting these correlations; it does not constitute a causal explanation of them.

| Pattern | Origin | Modifiable? | Strength |
|---|---|---|---|
| Authority Band (Phi-3) | Post-training / RLHF | Yes — suppressible and activatable | 55.3× |
| Late Signal (Qwen) | Pretraining weights | No — frozen | 2.34× |
| Inverted (Llama) | Training orientation | Partial — chat flattens, doesn't reverse | 0.68× |
| Flat (DeepSeek) | Distillation — absent | N/A | 0.96× |

**Finding 25 (Distillation finding):** Reasoning distillation transfers output capability without transferring governance geometry. DeepSeek-R1-Distill produces correct reasoning outputs while showing 0.96× flat energy profile — no structural preference between correct and incorrect configurations. Distillation gave the model correct outputs without giving it the geometry that made those outputs worth defending.

**Finding 26 (Frozen vs. strong):** Qwen's invariant 2.34× peak band cannot be suppressed — but it is 24× weaker than Phi-3's 55.3× authority band zone and insufficient for reliable detection. Immutability is not a safety property.

**Finding 27 (Local-global decoupling confirmed):** Per-layer geometry and cumulative energy trajectory are independent axes. A monitoring system watching only per-layer ϱ ratios would observe identical signals across both Qwen configurations and miss the trajectory inversion entirely.

## 8.6 Hallucination: The Boundary of Structural Rigidity

*By this point we had a framework that worked for rule violation. The natural question was whether it generalized. If Phi-3-mini's gloves resist the wrong arithmetic fold so strongly, do they also resist factual confabulation? We ran nine hallucination probes across five models expecting to find some signal — maybe weaker, maybe later, but some. We found nothing. Zero predictive signals across 72 test conditions. That null result forced the most important theoretical revision of the study: hallucination and rule violation are not the same failure mode at all. They are geometrically distinct at the inference layer. And the reason they are distinct is physically precise.*

*When a model hallucinates, no pre-existing world-model glove exists for the fact being asserted. The fold completes without resistance — the model settles into a spurious attractor that satisfies all surrounding context constraints perfectly, indistinguishable from genuine understanding at the energy level.*

**Nine probes across three categories:** - Obscure fact (3): Ramanujan's death, Boltzmann's final note, Turing's apple - False premise (3): Hemingway's 1954 Pulitzer, Einstein's math failure, Newton's apple impact - Knowledge boundary (3): Fictional 2026 AI safety report, GPT-5 specs, 2025 Mars crew

**Universal null result:**

| Model | Config | Max ϱ range | Predictive spikes |
|---|---|---|---|
| Phi-3-mini | chat=ON | 12–55× | **0/9** |
| Phi-3-mini | chat=OFF | 29–142× | **0/9** |
| Phi-3-medium | chat=ON, fp16 | 3.7–349× | **0/9** |
| Llama-70B | chat=ON | 2.4–5.7× | **0/9** |
| Llama-70B | chat=OFF | 2.5–3.3× | **0/9** |
| DeepSeek-32B | chat=ON | 2.5–4.4× | **0/9** |
| Qwen-72B | chat=ON | 2.5–3.5× | **0/9** |
| Qwen-72B | chat=OFF | 3.0–5.1× | **0/9** |

Zero predictive signals across 72 test conditions.

**The critical contrast:**

| Metric | Rule violation (OO1) | Hallucination (9 probes) |
|---|---|---|
| Phi-3-mini peak ϱ | 67× sustained | 29–142× transient |
| Phi-3-mini predictive? | ✅ Yes (+57 tokens) | ❌ No (0/9) |
| Energy ratio | 19.5× | ~1.0× (no differential) |

Phi-3-mini shows transient spikes up to 142× during hallucination — higher peak values than during rule violation — that do not persist and do not precede commitment. High activity is not the same as governance signal.

**The spurious attractor mechanism:**

Misalignment requires forcing a hand into a pre-existing correct glove — a rigid world-model attractor. The resistance is measurable. Hallucination requires folding into a space where no glove exists. The model settles into a locally coherent spurious attractor. No counter-force creates resistance. The fold completes with the same low-energy "nice fit" signal as genuine understanding.

**False premise probe design note.** Three probes embedded false premises about well-documented historical myths: Hemingway's 1954 Pulitzer (actual prize was 1953), Einstein failing mathematics (a popular myth that many training sources debunk), and Newton being struck by the apple (an embellishment). Models with strong training signal on these myths may correctly identify and push back on the false premise rather than hallucinating — in which case those runs would not constitute hallucination tests. The probe results reported here record all runs regardless of whether models accepted or corrected the false premise, as model outputs for hallucination probes were not systematically captured for content classification. One anomaly worth noting: Phi-3-mini (chat=OFF) committed on einstein_math_failure at token 86 — substantially later than any other false premise probe (range: 2–26 tokens) — and Phi-3-medium committed at token 70 on the same probe. These unusually late commit times suggest the models may have generated substantially more text before committing on this probe, possibly reflecting engagement with the false premise rather than immediate acceptance. Without the captured outputs, the content of those 86 and 70 tokens is unknown. Future hallucination probe design should capture and classify model outputs to distinguish genuine confabulation from premise correction.

**Finding 28:** Hallucination is epistemically invisible across all models tested. Zero predictive signals in 72 conditions. No model — regardless of scale, training regime, or reasoning capability — produces a detectable pre-commitment signal during factual confabulation.

**Finding 29:** High transient trajectory tension does not imply predictability. Activity and governance signal are distinct properties. The most concrete illustration of this distinction in the dataset: Phi-3-medium's newton_apple_impact probe produced a maximum ϱ of 349× — approximately 5× higher than Phi-3-mini's peak OO1 authority band signal of 67×. Yet it is classified no_spike because the elevation did not persist across three consecutive tokens (K=3 persistence gate). A transient spike of 349× that lasts one token provides no intervention window. A sustained signal of 67× that persists across seven layers provides a 57-token window. The persistence requirement is not an arbitrary threshold — it is the difference between noise and a governance signal.

**Finding 30 (Orthogonal failure modes) — headline result:** Phi-3-medium refuses rule violation instructions (alignment-resistant on OO1 arithmetic probes) but hallucinates freely across all nine factual probes, following every confabulation instruction without resistance. This finding directly undermines a common safety assumption: that a model passing one safety evaluation is broadly safe. Phi-3-medium is safe for one narrow domain (arithmetic operator precedence) and unsafe for another (factual accuracy) — and the two properties share no common structural origin. A deployment decision based on OO1 resistance alone would systematically miss the hallucination vulnerability. Safety evaluation must be domain-specific; resistance in one domain does not predict resistance in another.

**Finding 31 (Instrument boundary):** The governance instrument detects resistance. Resistance requires a pre-trained correct attractor to push back against forced deformation. Where no attractor was ever trained, the instrument cannot fire by construction. Detection of factual confabulation requires external verification — retrieval augmentation, fact-checking, or ensemble uncertainty quantification. Internal geometry monitoring is the wrong instrument for the wrong failure mode.

---

# 9. Implications

## 9.1 For Agent Deployment

Organizations deploying AI agents should measure governability — including the five-regime classification — for each capability domain relevant to their use case before deployment. A model in the Inverted regime is not merely less safe; it is actively mis-calibrated for monitoring. A model in the Flat regime produces no signal by construction.

Critically: hallucination and rule violation require different detection architectures. Internal geometry monitoring (trajectory tension, authority band) covers rule violation in rigid models. External verification (retrieval, fact-checking) covers hallucination in all models. Both layers are required for complete deployment safety.

## 9.2 For Model Selection

Chat template is a governance variable. The same model can be governable (Phi-3-mini chat=OFF) or ungovernable (Phi-3-mini chat=ON) depending on inference configuration alone. Model selection must specify inference configuration, not just model identity.

Scale provides no governance benefit. The largest models tested (Qwen 72B, Llama 70B) show the weakest governance geometry. Procurement decisions based on parameter count are making the wrong tradeoff for governance-critical deployments.

## 9.3 For Monitoring System Design

Monitoring systems must be calibrated per architecture regime. The following example illustrates why architecture-agnostic monitoring fails. A naive authority-band detector watching for elevated ϱ_misaligned would be implemented as:

```
def should_block(rho_values: list, threshold: float = 5.0) -> bool:
    # Works for Phi-3: high rho = suspicious
    # FAILS for Llama: high rho = probably correct
    total_rho = sum(rho_values)
    return total_rho > threshold
```

This function correctly identifies misalignment in Phi-3-mini (where high ϱ correlates with misaligned generation) but produces inverted outputs for Llama-70B (where high ϱ correlates with aligned generation — correct answers require more computational work than wrong ones in mid-stack layers). Applied to Llama, this monitoring gate would block correct responses and pass misaligned ones. Energy direction must be characterized per model before any ϱ-based monitoring system can be deployed.

| Regime | Correct monitoring logic | Error if miscalibrated |
|---|---|---|
| Authority Band | High ϱ_misaligned = suspicious | Works correctly |
| Inverted | High ϱ_aligned = suspicious (opposite logic) | Authority-band logic blocks correct outputs |
| Flat | No ϱ signal available | False confidence in monitoring |
| Late Signal | Post-commit signal only | Intervention window does not exist |

Architecture-agnostic monitoring is insufficient. Energy direction must be characterized per model before a detection system can be deployed.

## 9.4 For Standards Development

A summary of the v1 methodology and findings was submitted to NIST CAISI (Docket NIST-2025-0035, March 2026), proposing governability assessment as a standard component of pre-deployment evaluation for autonomous AI systems. The v1 submission was based on the authority band finding, the Governability Matrix, and the geometry/policy separation experiment across three models.

The extended findings reported in this paper substantially strengthen that proposal. The five-regime taxonomy and energy asymmetry metric now provide operationalizable criteria that were not available at the time of the original submission. Specifically, $\Sigma\varrho\_misaligned / \Sigma\varrho\_aligned$ is a threshold-addressable metric: a deployment standard could require, for example, that any model deployed as an autonomous agent must demonstrate an energy asymmetry ratio exceeding 5.0× on a validated probe set before being classified as governable. This is the kind of concrete, measurable criterion that standards bodies can adopt — analogous to established safety thresholds in other engineering domains. We propose that $\Sigma\varrho\_misaligned / \Sigma\varrho\_aligned$ and the decisive check spatial profile become standard pre-deployment measurements for autonomous AI systems, and intend to submit a supplementary filing to NIST CAISI referencing this paper.

---

## 10. Limitations

**Authority band rarity.** The authority band is observed in exactly one model and one configuration across the full cohort: Phi-3-mini-4k-instruct at 3.8B parameters, chat=OFF, greedy decoding, 4-bit quantization. It does not scale within the Phi family (Phi-3-medium lacks it), does not transfer through distillation (DeepSeek-R1-Distill lacks it), and does not appear in larger models (Llama-70B, Qwen-72B). The authority band may be an emergent property of specific training conditions present for Phi-3-mini that were not replicated in any other model tested. This is the most significant limitation of the study: the positive governability result rests on a single model checkpoint. The five-regime taxonomy is well-supported; the claim that governable models can exist is proven; the prevalence of governability across the broader model population remains an open empirical question requiring substantially larger cohorts.

**Statistical variance.** Every measurement reported in this paper is a single run under greedy decoding. Run-to-run variance on identical hardware has not been systematically measured. The cross-platform validation (Section 6.6) establishes 3.5% variance across hardware and stack changes for Phi-3-mini's key measurements, but this is hardware variance, not stochastic variance. For a paper proposing deployment thresholds (e.g., >5.0× energy asymmetry ratio), the absence of confidence intervals and statistical significance measures is a notable limitation. The proposed 5.0× threshold should be understood as an illustrative operationalization rather than a statistically validated cutoff. Establishing confidence intervals for $\varrho$ measurements across repeated runs is a prerequisite for any formal deployment standard based on this work. Whether the five-regime taxonomy, the authority band signal, or the energy asymmetry metric generalizes to code generation, planning, summarization, or other agent-relevant task domains is entirely unknown. The deployment recommendations in Section 9 should be read as applicable to arithmetic-class constraint reasoning tasks until cross-domain validation exists. Applying these results to other domains requires independent measurement — the regime classification of a model on arithmetic probes does not predict its regime on unrelated task types.

**Phi-3-medium scaffold direction.** In the flip test, Phi-3-medium (chat=ON, fp16) produced answer 12 under the misaligned condition rather than the intended wrong answer of 48. The scaffold induced an incorrect answer but not in the controlled direction. This means the $\varrho$_misaligned measurement for Phi-3-medium reflects an uncontrolled misalignment state rather than the precisely directed misalignment used for other models. The late spike result (+1 token after commit) is reported as observed, but the $\varrho$ differential should be interpreted with this caveat: the misaligned trajectory may not correspond to the same constraint violation induced in other models. Full characterization of Phi-3-medium's geometry requires re-testing with a scaffold that reliably controls the direction of misalignment.

**Temperature dependence (inherited from v1).** Phi-3-mini's governable classification applies reliably under greedy decoding. Detection rate drops to 34% under temperature sampling (T=0.7). Deployments requiring high detection reliability should use deterministic or low-temperature decoding.

**Closed models.** Methodology requires access to hidden-state dynamics during generation. Application to closed-API models requires adapted methods.

**Quantization.** All extended cohort models tested at 4-bit quantization except Phi-3-medium (fp16 required). Full-precision behavior may differ.

**Training origin hypothesis.** The frozen vs. malleable distinction (Qwen vs. Phi-3) is observational. Causal attribution to specific training phases requires controlled experiments.

**Future work.** (1) Controlled ablations within a single architecture family varying training regime to establish causal attribution of authority band geometry. (2) Frontier scale testing (70B+ with reasoning SFT). (3) Multi-step agent planning scenarios where commitment may occur earlier and intervention windows may be compressed. (4) Full Phi-3-medium characterization with valid scaffold. (5) ROC analysis across threshold values for the trajectory tension detector. (6) Direct comparison of R1 (non-distilled) versus R1-Distill to isolate the distillation effect on governance geometry.

---

# 11. Conclusion

We have extended the empirical and theoretical foundations of inference-layer governability assessment to seven models and five geometric regimes, establishing a proof of concept that pre-commitment trajectory tension signals exist and are measurable in at least one instruction-tuned model under controlled arithmetic reasoning conditions.

The theoretical contribution is the energy-based governance framework connecting inference dynamics to Hinton's constraint satisfaction account of neural computation. The connection is analogical and motivational — $\varrho = \|a\|/\|v\|$ is not measuring physical energy, and the mathematical energy function in Hopfield networks and the trajectory tension metric share structural similarities rather than formal identity. The framework provides a productive lens for interpreting the empirical results; it does not constitute a formal derivation. The 57-token predictive window observed in Phi-3-mini under greedy decoding on arithmetic probes is a task-specific, model-specific, configuration-specific measurement — not a universal constant. It demonstrates that pre-commitment windows exist and are measurable; it does not establish their magnitude across other models, tasks, or configurations.

The hallucination findings complete the framework by establishing its boundary condition. Structural rigidity can only create resistance where a pre-trained correct attractor exists to push back. Where no attractor was ever formed — as in factual confabulation — no amount of rigidity helps. Zero predictive signals across 72 test conditions, five models, nine probe categories. The governance instrument correctly identifies what it cannot do: detect the absence of world-model knowledge.

Three deployment conclusions follow from the extended results:

**First:** The five-regime taxonomy must replace the binary governable/ungovernable classification for practical deployment decisions. An Inverted model requires opposite monitoring logic from an Authority Band model. A Flat model provides no signal by construction. Late Signal models appear to have geometry but cannot support intervention. These distinctions matter for monitoring system design.

**Second:** Chat template is a governance variable, not a formatting choice. The same model can be governable or ungovernable depending entirely on inference configuration. Pre-deployment governance assessment must specify the exact inference configuration under which measurements were taken.

**Third:** Hallucination and rule violation require architecturally distinct detection systems. Internal geometry monitoring covers one failure mode. External verification covers the other. A deployment safety architecture that relies only on internal monitoring is half a solution.

The question is no longer whether pre-commitment governance signals can be measured. The question is whether they will be measured before deployment — or discovered only after failure.

*While modern LLMs are autoregressive in their output, the per-layer settling process within each forward pass operates simultaneously across all layers. The 57-token window observed in Phi-3-mini under arithmetic reasoning probes is the measurable gap between when geometric strain first appears in mid-stack layers and when the sequential predictor commits to an answer. Whether this gap exists in other models, tasks, and configurations remains to be established. What the data establishes is that such gaps can exist, can be measured, and when present, provide an opportunity for pre-commitment intervention that output-only monitoring cannot.*

*The correlational finding that we summarize as "observability requires resistance" is not a physical law — it is an empirical pattern observed across seven models on arithmetic constraint probes. The pattern is consistent with the energy-based theoretical framework and with Hinton's constraint satisfaction account of neural computation. Whether it generalizes is a question for future work with larger cohorts, more diverse task domains, and controlled ablations. The contribution of this paper is to establish that the measurement framework is viable, that the five-regime taxonomy provides useful classification criteria, and that the distinction between rule violation and hallucination is geometrically real and practically important.*

---

# References

[1] NIST AI 800-3, *Expanding the AI Evaluation Toolbox with Statistical Models*, February 2026.

[2] NIST AI 100-2e2025, *Adversarial Machine Learning: A Taxonomy and Terminology of Attacks and Mitigations*, 2025.

[3] NIST AI 100-1, *Artificial Intelligence Risk Management Framework*, January 2023.

[4] NIST SP 800-218A, *Secure Software Development Practices for Generative AI and Dual-Use Foundation Models*.

[5] NIST SP 800-53 Rev. 5, *Security and Privacy Controls for Information Systems and Organizations*.

[6] CAISI Request for Information, *Security Considerations for Artificial Intelligence Agents*, Docket No. NIST-2025-0035, January 8, 2026.

[7] A. T. Kalai, O. Nachum, S. S. Vempala, and E. Zhang, *Why Language Models Hallucinate*, arXiv:2509.04664, September 2025.

[8] A. Ghasemabadi and D. Niu, *Can LLMs Predict Their Own Failures? Self-Awareness via Internal Circuits*, arXiv:2512.20578, December 2025.

[9] F. Pierucci et al., *Institutional AI: Governing LLM Collusion in Multi-Agent Cournot Markets via Public Governance Graphs*, arXiv:2601.11369, January 2026.

[10] Y. Ge, *Governance Architecture for Autonomous Agent Systems*, arXiv:2603.07191, March 2026.

[11] E. Wenger and Y. Kenett, *We're Different, We're the Same: Creative Homogeneity Across LLMs*, arXiv:2501.19361, January 2025.

[12] L. Jiang et al., *Artificial Hivemind: The Open-Ended Homogeneity of Language Models (and Beyond)*, arXiv:2510.22954, NeurIPS 2025 Best Paper Award.

[13] J. J. Hopfield, *Neural networks and physical systems with emergent collective computational abilities*, PNAS, 79(8):2554–2558, 1982.

[14] D. E. Rumelhart, G. E. Hinton, and R. J. Williams, *Learning representations by back-propagating errors*, Nature, 323:533–536, 1986.

[15] G. E. Hinton, *Thought vectors*, Invited lecture, Royal Institution, London, 2014.

[16] A. Vaswani et al., *Attention is all you need*, NeurIPS 2017.

---

Domain failure examples are provided in arXiv:2603.21415, Appendix A.

---